\documentclass[10pt,twocolumn,letterpaper]{article}

\usepackage{cvpr}
\usepackage{times}
\usepackage{epsfig}
\usepackage{graphicx}
\usepackage{amsmath}
\usepackage{amssymb}
\usepackage{bm}

\usepackage{makecell}
\usepackage{soul}

\usepackage{commath}
\usepackage{caption}
\usepackage{wrapfig}
\usepackage{booktabs}       
\usepackage{amsfonts}       
\usepackage{nicefrac}       
\usepackage{microtype}      
\usepackage{mathtools}
\usepackage{cuted}
\usepackage{caption}
\usepackage{subcaption}
\usepackage{breqn}
\usepackage{sidecap}
\usepackage{float}
\usepackage{amsmath,bm}

\usepackage[ruled,vlined,noend]{algorithm2e}

\usepackage[pagebackref=true,breaklinks=true,letterpaper=true,colorlinks,bookmarks=false]{hyperref}

\newcommand{\best}[1]{\textcolor{red}{\textbf{#1}}}%
\newcommand{\sbest}[1]{\textcolor{blue}{\textbf{#1}}}%

\cvprfinalcopy 
\newcommand{\venue}[1]{{\scriptsize #1}}

\def\cvprPaperID{8831} 

\ifcvprfinal\fi
\begin{document}

\title{\vspace{-0.5em}Spectral-GANs for High-Resolution 3D Point-cloud Generation}

\author{Sameera Ramasinghe, Salman Khan, Nick Barnes, Stephen Gould\\
Australian National University\\
{\tt\small firstname.lastname@anu.edu.au}
}

\maketitle

\begin{abstract}\vspace{-0.5em}
Point-clouds are a popular choice for vision and graphics tasks due to their accurate shape description and direct acquisition from range-scanners. This demands the ability to synthesize and reconstruct 
high-quality 
point-clouds. Current deep generative models for 3D data generally work on simplified representations (e.g.,  voxelized objects) and cannot deal with the inherent redundancy and irregularity in point-clouds.  A few recent efforts on 3D point-cloud generation offer limited resolution and their complexity grows with the increase in output resolution. In this paper, we develop a principled approach to synthesize 3D point-clouds using a spectral-domain Generative Adversarial Network (GAN). Our spectral representation is highly structured and allows us to disentangle various frequency bands such that the learning task is simplified for a GAN model. 
As compared to spatial-domain generative approaches, our formulation allows us to generate 
high-resolution point-clouds with minimal computational overhead. Furthermore, we propose a fully differentiable block to transform from  {the} spectral to the spatial domain and back, thereby allowing us to integrate knowledge from well-established spatial models.  We demonstrate that Spectral-GAN performs well for point-cloud generation task. Additionally, it can learn  {a} highly discriminative representation in an unsupervised fashion and can be used to accurately reconstruct 3D objects.  
\end{abstract}


\begin{figure}[!tp]
    \centering
    \includegraphics[width=\columnwidth]{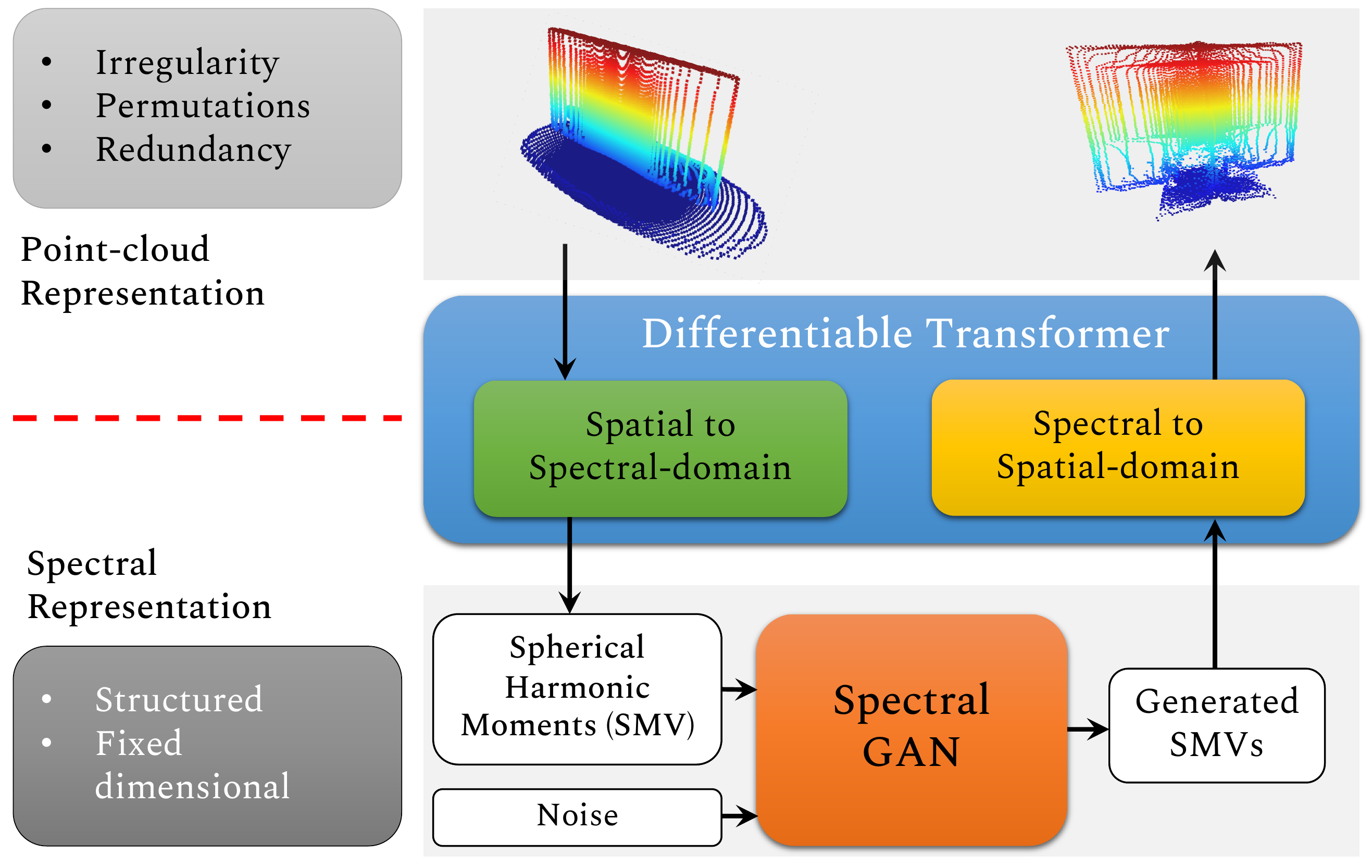}\vspace{-0.8em}
    \caption{\small\emph{Overview of Spectral-GAN.} Our model operates in the spectral domain using spherical harmonic moment vectors (SMVs). This allows us to avoid the redundancy and irregularity of point-clouds. Using a differentiable transformer, our model can also receive guidance from the spatial domain.
    }
    \label{fig:my_label}
\end{figure}

\vspace{-0.8em}
\section{Introduction}\vspace{-0.3em}

Point-clouds are a popular 3D representation for real-world objects and scenes. In comparison to other representations such as voxels, mesh and truncated signed distance function (TSDF),  point-clouds are often an attractive choice for 3D data because they capture shape details accurately, are computationally efficient to process and can be acquired as a default output from several 3D sensors (e.g., LiDAR). However, point-clouds pose a major challenge for deep networks, particularly the generative pipelines, due to their inherent redundancy and irregular nature (e.g., permutation-invariance).

Due to the complexity of point-clouds, most 3D synthesis approaches are inapplicable. For example, generative approaches designed for voxelized inputs \cite{wu2016learning, kingma2013auto, wu20153d, xie2018learning, Huang_2017_CVPR, khan2019unsupervised},  cannot work with the irregular point sets. To overcome this challenge, some recent generative approaches solely focus on point-cloud synthesis.  For example, Achlioptas \etal \cite{achlioptas2017learning} use a GAN framework for 3D point-cloud distribution modelling in the data and auto-encoder latent space, Yang \etal \cite{Yang_2019_ICCV} sample 3D points from a prior spatial distribution and then transform them using an invertible parameterization while \cite{shu20193d, valsesia2018learning} employ graph-structured 
networks for point-cloud generation.  

All such efforts so far, operate in the `spatial-domain' (3D Euclidean space) which makes the modelling task relatively difficult due to arbitrary point configurations in 3D space. This leads to a number of roadblocks towards a versatile generative model e.g., considering a fixed set of points  \cite{achlioptas2017learning}  
and limited scalability to arbitrary point resolutions \cite{shu20193d, valsesia2018learning}. As opposed to previous works, we perform generative modelling in the spectral space using spherical harmonic moment vectors (SMVs), which
inherently offers a solution to the above mentioned problems. 
Specifically, generating 3D shapes via spectral representations allows us to compactly represent redundant information in point-clouds, easily scale to high-dimensional point-cloud sets, remain invariant to the permutations in unordered point sets and generate high-fidelity shapes with relatively minimal outliers. Besides, our spectral representation allow us to develop an understanding about the frequency domain functional space of generic 3D objects. Our main contributions are:\vspace{-0.5em}
\begin{itemize}\setlength{\itemsep}{-0.3em}
\item To handle the redundancy and irregularity of point-clouds, we propose the first spectral-domain GAN that  synthesizes novel 3D shapes by using a spherical harmonics based representation. 
\item A fully differentiable transformation from the spectral to the spatial domain and back, thus allowing us to integrate knowledge from well-established spatial models.
\item Through both quantitative and qualitative evaluations, we illustrate that Spectral-GAN can generate high-quality 3D shapes with minimal artifacts and can be easily scaled to high-dimensional outputs.
\item Our proposed framework learns highly discriminative unsupervised  features and can seamlessly perform 3D reconstruction from 2D inputs. Moreover, we show that Spectral-GAN is scalable to high-resolution outputs (40$\times$ resolution increase with just 4$\times$ parameters). 
\end{itemize}

\section{Related Work}


\textbf{Generative models in spectral-domain:} Yang \etal \cite{yang2017dagan} and Souza \etal \cite{souza2018hybrid} develop methods for MRI reconstruction using GANs, and use Fourier domain information to refine the output. In the former approach, the generator operates in the spatial domain, and 
spectral information is used to refine the output. The latter approach, in contrast, uses two separate networks in the frequency and spatial domains and adopts the Fourier transform to exchange information between the two. A significant drawback of these approaches is that output resolution  is tightly coupled to the network design and thus they lack scalability to high dimensions.


In a different application, Portilla \textit{et al.} \cite{portilla2000parametric} present a method to synthesize textures as 2D images based on a complex wavelet transform. They parameterize this operation using a set of statistics computed on pairs of coefficients corresponding to basis functions at adjacent spatial locations, orientations, and scales. However, their approach is not a learning model, which offers less flexibility. 
Furthermore, Zhu \textit{et al}. \cite{zhu2018image} recently proposed a model that initially processes undersampled input data in the frequency domain and then refines the result in the spatial domain using the inverse Fourier transform. They approximate the inverse Fourier transform using a sequence of connected layers, 
but one disadvantage  is that their transformation  has quadratic complexity with respect to the size of the input image. Furthermore, the above works are limited to 2D and do not study the 3D point-cloud generation problem in spectral domain.

\textbf{3D GANs in spatial-domain:} 3D GANs can be primarily categorized into two types: voxel outputs and point-cloud outputs. The latter typically entails more challenges as point-clouds are unordered and highly irregular in nature. 

For voxelized 3D object modeling, several influential methods have been proposed in the literature. Wu \textit{et al.} \cite{wu2016learning} 
extend the 2D GAN framework to 3D domain for the first time. 
Following their work, Smith \textit{et al.} \cite{smith2017improved} use a novel GAN architecture for 3D shape generation by employing Wasserstein distance as the loss function. 
A recent work by Khan \textit{et al.} \cite{khan2019unsupervised} presents a  factorized 3D generative model that sequentially generates shapes in a coarse-to-fine manner. 
Our approach also uses a two-step procedure--a forward pass and backward pass---to refine a coarse 3D shape, but a key difference here is that they use spatial information to refine the shape, while our method depends on frequency information. 

Naive extensions of traditional spatial GANs to 3D point-cloud generation do not produce satisfactory results, due to their inherent properties such as being an unordered, irregularly distributed collection (see Sec.~\ref{sec:problem}). 
Achlioptas \etal \cite{achlioptas2017learning} were the first to use GANs to generate point-clouds. 
They first convert a point-cloud to a compact latent representation and then train a discriminator on it. Although we also use a compact representation, \ie, the SMV to train the GAN, SMVs provide a richer representation compared to latent space approximations and theoretically guarantee accurate reconstruction of the 3D point-cloud. Moreover, Valsesia  \etal \cite{valsesia2018learning} propose a graph convolution based network to extract localized features from 3D point-clouds, in order to reduce the effect of irregularity. A drawback of their method, however, is the rather high computational complexity of graph convolution, and less scalability with the resolution of the point-cloud. A recent work by Shu \etal \cite{shu20193d} also propose a tree-structured graph convolution network, which is more computationally efficient. 
The model proposed by Li \etal \cite{li2018point} attempts to handle the irregularity of point-clouds using a separate inference model which captures a latent distribution, 
to deal with the irregularity of point-clouds. 
In contrast, we effectively reduce the problem to the standard GAN setting by using a fixed-dimensional 
representation for point-clouds.



\section{Problem Formulation}
\label{sec:problem}
 
 An \textit{exchangeable} sequence can be considered as a sequence of random variables $\Tilde{X} = \{x_i\}_{i=1}^{n}$, where the joint probability distribution of $\Tilde{X}$ does not vary under position permutations. More formally,
 
 \noindent \textbf{Definition: } \textit{For a given finite set $\{x_i\}_{i=1}^{n}$ of random objects, let $\mu_{x_1, x_2, \hdots, x_n}$ be their joint distribution. This finite set is exchangeable if $\mu_{x_1, x_2, \hdots, x_n} = \mu_{x_{\pi(1)}, x_{\pi(2)}, \hdots, x_{\pi(n)}}$, for every permutation $\pi: \{1,2,\hdots, n \} \rightarrow \{1,2,\hdots, n \}$.}
 The spatial representation $X$ of a point-cloud is a \textit{set} of $d$-dimensional vectors, and in cases of Euclidean geometry, typically, $d=3$. A \textit{set} is a collection of elements without any particular order or a fixed number of elements and thus, the probability distribution $p(X)$ is an exchangeable sequence. According to the Hewitt and Savage theorem \cite{hewitt1955symmetric}, there exists a latent distribution $Q$ such that,
 \begin{align}
     \label{equ:hewatt}
     p(X) = \int p(x_1, x_2, \hdots, x_n \mid Q)\, p(Q)dQ.
 \end{align}
Eq.~\ref{equ:hewatt} shows that in order to properly model $X$ as an exchangeable sequence and obtain a distribution $p(X)$, it is necessary to capture the  latent representation $Q$. In other words, it is difficult for a GAN to model $X$ as an exchangeable sequence, only by observing a set of $X$ sequences and estimating the marginal distributions $p(x_i), i \in [1, \ldots, n]$. In this case, the generative model needs to learn the joint probability distribution $p(x_i,Q)$ instead of $p(x_i)$. 
This makes it challenging to extend traditional GANs to the point-cloud generation problem. A straightforward approach to resolve this is to model point-cloud data as ordered, fixed-dimensional vectors. However, this approach does not hold the integral probability metric (IPM) guarantees of a GAN \cite{li2018point}. 

On the contrary, we propose to model point-cloud data as SMVs, which effectively reduces the problem to the traditional case in two ways: 1) SMVs encode the corresponding shape information in a structured, fixed dimensional vector and 2) the vector elements are highly correlated with each other. The task of learning the distribution of elements of SMVs is theoretically similar to learning the pixel distribution of images, as in the latter case also, we only need to capture the joint probability distribution of pixels of each instance. In the case of image synthesis, however, GANs exploit the correlation of pixels using convolution kernels, which is not possible in the case of SMVs as correlation does not depend on proximity. Furthermore, different frequency portions of the SMVs show different characteristics. To handle these specific attributes, we propose a series of cascaded GANs, each consisting of only fully connected layers. Since each GAN only needs to generate a specific portion of the SMV, they can be designed as shallow models with fewer floating point operations (FLOPs).

\section{Spectral GAN}
We propose a 3D generative model that operates entirely in the spectral domain. Such a design offers unique advantages over spatial domain 3D generative models: (a) a compact representation of 3D shapes with an intuitive frequency-domain interpretation, (b) the flexibility to generate high-dimensional shapes with minimal changes to the model complexity, and (c) a permutation invariant representation which handles the irregularity of point-clouds.  Below, we first introduce the spherical harmonics representations that serve as the basis for our proposed Spectral GAN model. 

\begin{figure*}[t!]
\centering
\includegraphics[width=0.9\textwidth]{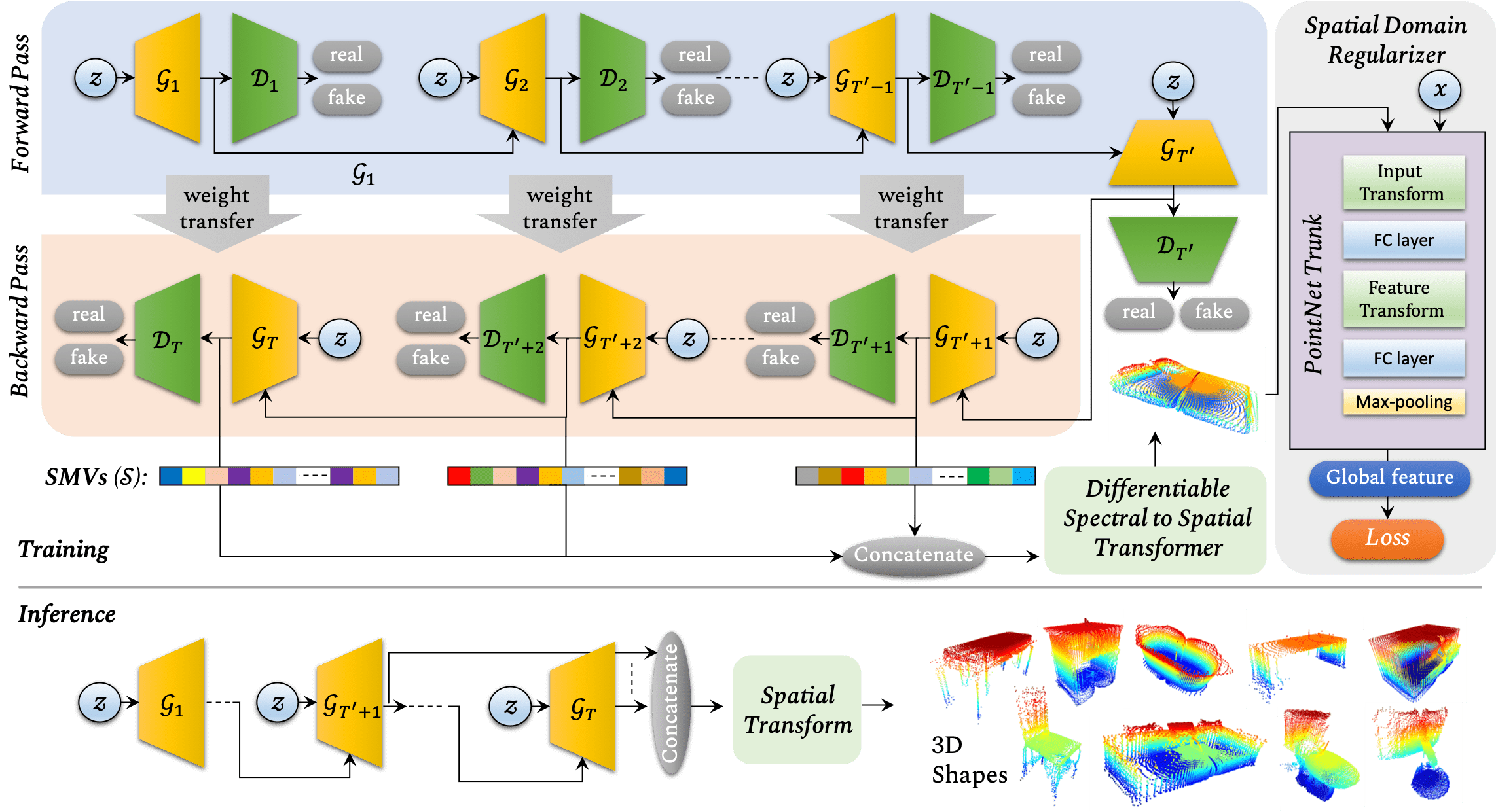}
\vspace{-1.2em}
\caption{\small The overview of the Spectral Generative Adversarial Network. An unrolled version (with an explicit forward and backward pass) of the training scheme is shown for clarity. }
\label{fig:overall}
\end{figure*}

\subsection{Spherical Harmonics for 3D Objects}
\label{sec:SMV}
Spherical harmonics are a set of complete and orthogonal basis functions, which can efficiently represent functions on the unit sphere $\mathbb{S}^2$ in $\mathbb{R}^3$. They are a higher dimensional analogy of the Fourier series, which forms a basis for functions on unit circle. The spherical harmonics are defined on $\mathbb{S}^2$ as, 
\begin{align}
Y^m_l (\theta, \phi) = N_l^m P_l^m(\cos\phi)e^{im\theta},
\end{align}
where $\phi \in [0,\pi]$ is the polar angle, $\theta \in [0, 2\pi ]$ is the azimuth angle, $l \in \mathbb{Z}^{+}$ is a non-negative integer, $m  \in \mathbb{Z}$ is an integer, $|m| < l$, $i=\sqrt{-1}$ is the imaginary unit, $N_l^m = (-1)^m\sqrt{\frac{2l+1}{4\pi}\frac{(l-m)!}{(l+m)!}}$ is the normalization coefficient and $P_l^m(\cdot)$ is the associated Legendre function,
\begin{align}
P_l^m(x) = (-1)^m \frac{(1-x^2)^{\frac{m}{2}}}{2^ll!}\frac{d^{l+m}}{dx^{l+m}}(x^2-1)^l.
\end{align}
Since spherical harmonics are orthogonal and complete over the continuous functions on $\mathbb{S}^2$ with finite energy, such a function $f:\mathbb{S}^2 {\rightarrow} \mathbb{R}$ can be expanded as,
\begin{align}\label{equ:expansion}
f(\theta,\phi) = \sum_{l=0}^{\infty}\sum_{m=-l}^{l}c^m_l Y^m_l(\theta,\phi), \quad
\end{align}
where $c^m_l$ are the spherical harmonic moments  obtained by,
\begin{align}\label{equ:moments}
    c^m_l = \int_0^{\pi} \int_0^{2 \pi}f(\theta, \phi) Y_{l}^{m}(\theta, \phi)^\dagger \sin\phi \, d\phi d\theta.
\end{align}
The sufficient conditions for the expansion in Eq.~\ref{equ:expansion} are given in \cite{hobson1931theory}. In practical cases, a bounded set of spherical harmonic basis functions $(M+1)^2$ is defined, where $M$ is the maximum degree of harmonics series.

The process of 3D shape modeling via spherical harmonics can be decomposed into two major steps. First, sample points from the 3D shape surface and then computing spherical harmonic moments. 
Any polar 3D surface function can be represented as $r = f(\theta, \phi)$, where $f(\theta, \phi)$ is a single valued function on the unit sphere $\mathbb{S}^2$, $r$ is the radial coordinate with respect to a predefined origin inside an object, and $(\theta, \phi)$ is the direction vector. Thus, we can compute moments of the corresponding 3D point-cloud using Eq. \ref{equ:moments}.




\subsection{Cascaded GAN Structure}
SMVs provide a highly structured representation of 3D objects, as explained in Sec.~\ref{sec:SMV}. Due to this structured nature, the margin for error is significantly lower in our setup, compared to GANs that try to produce spatial domain representations. Also, different frequency bands of the SMV typically entail different characteristics, which makes it highly challenging for a single GAN to generalize over the complete SMV. Therefore, to overcome this obstacle, we use multiple cascaded GANs, where each GAN specializes in generating a pre-defined frequency band of the SMV.

Our approach uses a combination of $T$ GAN models to generate the SMV of 3D shapes. Among them, the first model is a regular GAN while the remaining $T-1$ models are conditional GANs (cGAN). The objective of initial GAN model is given by a two-player min-max game,
\begin{align}
    \min_{\mathcal{G}_1} \max_{\mathcal{D}_1} L_{GAN}(&\mathcal{G}_1, \mathcal{D}_1) =    \mathbb{E}_{\bar{\bm{g}}_1} [\log \mathcal{D}(\bar{\bm{g}}_1)] + \notag\\
    & \mathbb{E}_{z_1}[\log (1- \mathcal{D}(\mathcal{G}(z_1)))],
\end{align}
where $\bar{\bm{g}}_i \sim p_{\bm{g}}$ is the SMV band sampled from the spectral coefficient distribution and $z \sim p_z$ is the noise vector sampled from a Gaussian distribution. In a cGAN, synthetic data modes are controlled by forwarding conditioning variables (e.g., a class label) as additional information to the generator. In our case, we use a specific band of SMVs $\bm{g}_i$ predicted by the previous generator to condition the subsequent generator. Then, the cGAN objective becomes,
\begin{align}
    \min_{\mathcal{G}_i} &  \max_{\mathcal{D}_i} L_{cGAN} (\mathcal{G}_i, \mathcal{D}_i)  =  \mathbb{E}_{\bar{\bm{g}}_i} [\log \mathcal{D}(\bar{\bm{g}}_i)]  + \notag\\
    & \mathbb{E}_{\bm{g}_{i-1}, z_i}[\log (1- \mathcal{D}(\mathcal{G}_i(\bm{g}_{i-1}, z_i)))] : i  > 1.
\end{align}

Each GAN generates a portion of the complete spherical moment vector for the next GAN to be conditioned upon. The setup includes two major steps: (i) forward pass and (ii) backward pass. Accordingly, the overall architecture can be decomposed into two sets of generators $\mathcal{G}_f$ and $\mathcal{G}_b$, that implement the forward and backward functions, respectively. In the forward pass, the model tries to generate a coarse shape representation, and the backward pass refines the coarse representation to generate a refined representation. The basis of our design is the \emph{Markovian assumption}, i.e., given the outputs from the neighbouring generators, a current generator is independent from the outputs of the rest. 
We describe the two generation steps  in Sec.~\ref{sec:forward} and \ref{sec:backward}. 

\subsubsection{Forward pass}
\label{sec:forward}
In the forward pass, we have a set of $T'$ generative models  $\mathcal{G}_f = \{\mathcal{G}_1,\hdots , \mathcal{G}_{T'}\}$, which work in unison to generate a coarse representation of a 3D shape. Each $\mathcal{G}_i \in \{\mathcal{G}_2, \hdots \mathcal{G}_{T'}\}$ is conditioned upon the outputs of $\mathcal{G}_{i-1}$, and generates a predefined frequency band ($\mathcal{S}_i$) of the complete spherical harmonic representation ($\mathcal{S}$) of the corresponding 3D shape. It is worthwhile to note that the forward pass is sufficient to generate the complete SMV without the aid of a backward pass. However, a critical limitation of this setup is that each GAN is only conditioned upon lower frequency bands of the SMV. In practice, this results in noisy outputs. Therefore, we also perform a backward pass, which allows the GANs to refine the generation by observing the higher frequencies. This procedure is explained on Sec.~\ref{sec:backward}.

\subsubsection{Backward pass}
\label{sec:backward}
As explained in Sec.~ \ref{sec:forward}, the aim of the backward pass is to generate a more refined SMV, which produces a more refined 3D shape. Similar to forward pass, the backward pass is implemented using another set of generators $\mathcal{G}_b = \{\mathcal{G}_{T'+1}, \hdots \mathcal{G}_{T}\}$, where $T = 2T'-1$. Each $\mathcal{G}_i \in \mathcal{G}_b$ is conditioned upon the outputs of $\mathcal{G}_{i-1}$ and generates a specific portion of the complete SMV. In the training phase, we first transfer the trained weights from $\{\mathcal{G}_f \backslash \mathcal{G}_{T'}\}$ to $\mathcal{G}_b$, before training $\{\mathcal{G}_b\}$. Therefore, this can be intuitively considered as fine-tuning $\{\mathcal{G}_1 \ldots \mathcal{G}_{T'-1}\}$ based on higher frequencies. The training procedure is explained in Sec.~\ref{sec:training}.


\section{Spatial domain regularizer}
Since SMVs are highly structured, each element of a particular SMV is crucial for accurate reconstruction of its corresponding 3D point-cloud. In other words, even slight variations of a particular SMV cause significant variations in the spatial domain. Therefore, it is cumbersome for a GAN to synthesize SMVs, corresponding to visually pleasing point-clouds, by solely observing a distribution of ground truth SMVs.

To surmount this barrier, we use a spatial domain regularizer that can refine the weights of our cascaded GAN architecture, in order to synthesize more plausible SMVs. The spatial domain regularizer provides feedback from the spatial domain to the GANs, depending on the quality of the spatial reconstruction. Firstly, we employ a  pre-trained  PointNet \cite{qi2017pointnet} model  on the reconstructed synthetic point-cloud, and extract a global feature. Secondly, using the same procedure, we obtain another global feature from a ground truth point-cloud from the same class, and compute the $L_2$ distance between these two features. Finally, using back back-propagation, we update the weights of all the generators $\bm{G}= \{\mathcal{G}_f \cup \mathcal{G}_b\}$ to minimize the $L_2$ distance. The final architecture of the proposed model is shown in Fig.~\ref{fig:overall}.

In order to back-propagate error signals from the spatial domain to the spectral domain, we require ${\partial \mathcal{L}}/{\partial \bm{g}}$, where $\bm{g}$ is the SMV and $\mathcal{L}$ is the loss. To this end, we derive the following formula: let $\bm{g} = (g_0^0, \hdots, g_l^m, \hdots g_K^K)^\top$ be the SMV of a particular instance and $\{r(\theta_0, \phi_0), \hdots, r(\theta_n, \phi_n), \hdots, r(\theta_N, \phi_N) \}$ be the corresponding reconstructed points on $\mathbb{S}^2$ for the same instance. Then, using the chain rule it can be shown that,
\begin{align}
    \frac{\partial \mathcal{L}}{\partial g_l^m} &= \sum_{\theta} \sum_{\phi} \frac{\partial \mathcal{L}}{\partial r(\theta, \phi)} \frac{\partial r(\theta, \phi)}{\partial g_l^m}, \label{equ:grad1} 
 \end{align}   \vspace{-1em}
\begin{align}
    \text{where, } & r(\theta, \phi) = \sum_{l=0}^{M} \sum_{m=-l}^{l} g_l^m Y_l^m (\theta, \phi). \label{equ:grad2}
\end{align}
Combining Eq. \ref{equ:grad1} and \ref{equ:grad2}, we obtain, 
\begin{align}
    \frac{\partial \mathcal{L}}{\partial g_l^m} =  \sum_{\theta} \sum_{\phi} \frac{\partial \mathcal{L}}{\partial r(\theta, \phi)} Y_l^m (\theta, \phi).
\end{align}
The above expression can be written as a matrix-vector product to obtain derivatives ${\partial\mathcal{L}}/{\partial \bm{g}}$. This makes the transformer a fully differentiable and a network-agnostic module which can be used to communicate between spectral and spatial domains. 



\section{Network architecture and training}
\label{sec:training}
 Our aim is to generate a compact spectral representation, \ie, a vector, instead of a irregular point set. 
 In the spatial domain, points are correlated across the spatial space, and convolutions can be adopted to capture these dependencies. In fact, convolution kernels extract local features, under the assumption that spatially closer data points form useful local features. In contrast, closer elements in spectral domain representations do not necessarily exhibit strong correlations. Therefore, convolutional layers fail to excel in this scenario and thus, we opt for fully connected (FC) layers in designing our GANs. Interestingly, however, our GANs learn to generate quality outputs with a low depth architecture. 

\textbf{Generator architecture:} For our main experiments, we choose the maximum degree of SMVs and the number of GANs as $M {=} 100$ and $T {=} 7$, respectively, where  $\mathcal{G}_f = \{\mathcal{G}_1, .., \mathcal{G}_4\}$ and $\mathcal{G}_b = \{\mathcal{G}_5, \mathcal{G}_6, \mathcal{G}_7\}$. Each generator in $\mathcal{G}_f$ respectively generates frequency bands ($0 \leq l \leq 50, -l \leq m \leq 0$), ($0 \leq l \leq 50, 0 < m \leq l$), ($50 < l \leq 100, -l \leq m \leq 0$) and ($50 < l \leq 100, 0 < m \leq l$). 
Since $\mathcal{G}_5, \mathcal{G}_6, \mathcal{G}_7$ are used to fine tune $\mathcal{G}_1, \mathcal{G}_2, \mathcal{G}_3$, they generate the same frequency portions as the latter set. 
For all the generators, we use the same architecture, except for the last FC layer. Each generator consists of three FC layers, first two layers with 512 neurons each, and the number of neurons in the last layer depends on the output size. For the first two layers, we use ReLU activation and the final layer has a $tanh$ activation.

\textbf{Training:} The input to each of our generators, except to $\mathcal{G}_1$, is a $300$-d vector: a $200$-d noise vector concatenated with a $100$-d vector sampled in equal intervals from the previous generator output. For $\mathcal{G}_1$, we use a $200$-d noise input. We use RMSprop as the optimization algorithm with $\rho {=} 0.9, \text{ momentum} {=} 0, \epsilon {=} 10^{-7}$, where symbols refer to usual notation. For $\mathcal{G}_f$ and $\mathcal{G}_b$, we use learning rates $0.001$ and $0.0001$ respectively, and for discriminators, we use a learning rate $10^{-5}$. While training, we use three discriminator updates per each  generator update. Our sampling procedure is explained in supplementary materials and the training scheme is illustrated in Algorithm~\ref{alg:train}.

\begin{algorithm}[t]
\SetAlgoLined
\small
$\bm{G}= \{\mathcal{G}_f \cup \mathcal{G}_b\}$;

$R_o$ = A set of samples from ground truth point-clouds;

\For{i iterations}{
  \For{each $\mathcal{G}_k \in \mathcal{G}_f$}{
  \For{j iterations}{
   Train $\mathcal{G}_k$\;
  }
  }
  $\mathcal{G}_b \xleftarrow[]{\textrm{Weights}}\{\mathcal{G}_1, \hdots \mathcal{G}_{T'-1} \}$\;
  
  \For{each $\mathcal{G}_k \in \mathcal{G}_b$}{
  \For{j iterations}{
   Train $\mathcal{G}_k$\;
  }
}

  \For{ p iterations }{
      $\bm{g} \xleftarrow[]{\textsc{Synthesize}}\{\mathcal{G}_{T'} \cup \mathcal{G}_b \}$\;
    
     $\bm{r}_g \leftarrow \textsc{Reconstruct}(\bm{g})$\;
    
     $f_g \leftarrow \textsc{PointNet}(r_g)$\; 
    
     $f_o \leftarrow \textsc{PointNet}(r_o \sim R_o)$\;
     
     $L \leftarrow \norm{f_g -f_o}_2$\;
     $\bm{G} \leftarrow \textsc{Update}(\bm{G},L)$\;
  }
 }
\caption{\small Training procedure for the Spectral-GAN.}
\label{alg:train}
\end{algorithm}


    
    
    
     
     


\section{3D reconstruction from single image}
\label{sec:reconstruction}

As a different application, we propose a generative model which can reconstruct 3D objects by observing a single RGB image. The model follows the network architecture proposed in Sec. \ref{sec:training}, with a few alterations. Instead of randomly choosing the latent vector $z$, we use a set of image encoders to obtain an object representative vector $\hat{z}$, by taking a 2D image as the input. We use the same image encoder proposed in \cite{wu20153d}, which consists of five spatial convolution layers with kernel size $\{11,5,5,5,8\}$ with strides $\{4,2,2,2,1\}$. We use batch normalization after each layer, and ReLu activation as the non-linearity. 

We use $T'$ such image encoders for each $\mathcal{G}_i \in \mathcal{G}_f$, and use the same $\hat{z}$ vectors generated for $\{\mathcal{G}_1, \hdots, \mathcal{G}_{T'-1}\}$ when training $\mathcal{G}_i \in \mathcal{G}_b$. Each image encoder is trained end-to-end with $\mathcal{G}_i \in \mathcal{G}_f$. The training procedure is similar to Algorithm \ref{alg:train}, although we use different loss functions in this case. To optimize the GANs in spectral domain, we use two loss components: an adversarial loss $\mathcal{L}_{ad}$ and a spectral reconstruction loss $\mathcal{L}_{sr}$. The final spectral domain loss $L_{spectral}$ is,
\begin{align}
    L_{spectral} = L_{ad} +\alpha L_{sr},
\end{align}
where $L_{sr}$ is the $L_2$ distance between the ground-truth SMV and the generated SMV from $\mathcal{G}_T' \cup \mathcal{G}_b $ and $\mathcal{L}_{ad}$ is given as,
\begin{align}
    \mathcal{L}_{ad} = \log \mathcal{D}(x) +  \log(1-\mathcal{D}(\mathcal{G}(\mathcal{E}(y))))
\end{align}
Here, $\mathcal{E}(\cdot)$ is the encoder function, $\mathcal{D}(\cdot)$, $\mathcal{G}(\cdot)$ and $y$ are discriminator function, generator function and image input, respectively. $\alpha$  is a scalar weight. For the spatial domain optimization, we replace spatial regularization loss with the Chamfer distance as follows:
\begin{align}
\label{equ:chamfer}
   L_{spatial} =  \sum_{u \in S_1} \min_{v \in S_2} \norm{u-v}_2^2 +  \sum_{v \in S_2} \min_{u \in S_1} \norm{u-v}_2^2
\end{align}
where $S_1$ and $S_2$ are ground-truth and generated point sets, respectively. First, we obtain $S_2$ by converting the SMV to a point-cloud using Eq.~\ref{equ:expansion} and then compute the loss (Eq.~\ref{equ:chamfer}). 


\begin{figure*}[ht!]
\centering
\includegraphics[width=1.0\textwidth]{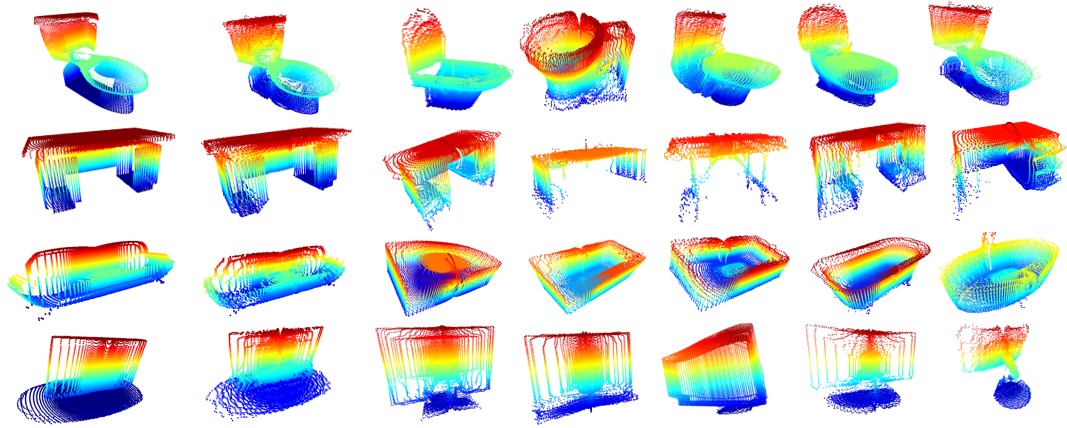}
\caption{\small Qualitative analysis of the results. From the left, $1^{st}$ column: Ground truth, $2^{nd}$ column: ground truth point-clouds reconstructed by SMV, $3^{rd} - 7^{th}$ columns: generated samples using spectral GAN.}
\label{fig:quantitative}
\end{figure*}

\section{Experiments}

In this section, we evaluate our model both qualitatively and quantitatively, and develop useful insights.


\subsection{3D shape generation}

\noindent \textbf{Qualitative results: } 
We train our model for each category in ModelNet10 and show samples of generated 3D point-clouds in Fig. \ref{fig:quantitative}. As expected, the reconstruction from SMV adds some noise to the ground truth point-clouds. An interesting observation, however, is that the quality of generated point-clouds are not far from from the reconstructed point-clouds from the ground-truth. Since the network only consumes the reconstructed ground-truth, this observation highlights the ability of our network in accurate modeling of input data distributions. 



\begin{table}
  \caption{\small 3D shape classification results on ModelNet10.}\vspace{-1em}
  \label{table:classification}
  \centering
  \scalebox{0.75}{
  \begin{tabular}{lcc}
    \toprule
    Method     & Type & Accuracy \\
    \midrule
    3D-ShapeNet \venue{(CVPR'15)} \cite{wu20153d} & Supervised & 93.5\% \\
    EC-CNNs \venue{(CVPR'17)} \cite{simonovsky2017dynamic} & Supervised & 90.0\% \\
    Kd-Network \venue{(ICCV'17)} \cite{klokov2017escape} & Supervised & 93.5\% \\
    LightNet \venue{(3DOR'17)} \cite{zhi2017lightnet} & Supervised & 93.4\% \\
    SO-Net \venue{(CVPR'18)} \cite{li2018so} &Supervised& 95.5\% \\
     \midrule
     Light Filed Descriptor \cite{chen2003visual} & Unsupervised & 79.9\% \\
     Vconv-DAE \venue{(ECCV'16)}  \cite{sharma2016vconv} & Unsupervised & 80.5\% \\
     3D-GAN \venue{(NIPS'16)} \cite{wu2016learning} & Unsupervised & 91.0\% \\
     3D-DesNet \venue{(CVPR'18)} \cite{xie2018learning} & Unsupervised & \sbest{92.4\%} \\
     3D-WINN \venue{(AAAI'19)} \cite{huang20193d} & Unsupervised & 91.9\%  \\
     PrimtiveGAN \venue{(CVPR'19)} \cite{khan2019unsupervised} & Unsupervised & 92.2\% \\
     \midrule
     Spectral-GAN (ours) & Unsupervised & \best{93.1\%}\\
    \bottomrule
  \end{tabular}}
\end{table}

\begin{SCtable}[][htp]
  \caption{\small Inception scores (IS) for 3D shape generation. We only compare with voxel based methods since no point-cloud (p-cloud) based method reports IS.}\hspace{-0.2cm}
  \label{table:inception}
  \centering\setlength{\tabcolsep}{2pt}
  \scalebox{0.7}{
  \begin{tabular}{lcc}
    \toprule
    Method     & 3D Data & Accuracy  \\
    \midrule
    3D-ShapeNet \cite{wu20153d} \venue{(CVPR'15)}&  voxel & 4.13 $\pm$ 0.19 \\
    3D-VAE \cite{kingma2013auto} \venue{(ICLR'15)}&  voxel  & 11.02 $\pm$ 0.42 \\
    3D-GAN \cite{wu2016learning} \venue{(NIPS'16)} &  voxel & 8.66 $\pm$ 0.45 \\
    3D-DesNet \cite{xie2018learning} \venue{(CVPR'18)} &  voxel & \best{11.77 $\pm$ 0.42} \\
    3D-WINN \cite{huang20193d} \venue{(AAAI'19)}&  voxel  & 8.81 $\pm$ 0.18 \\
    PrimitiveGAN \cite{khan2019unsupervised} \venue{(CVPR'19)}  & voxel & 11.52 $\pm$ 0.33 \\
    \midrule
    Spectral-GAN (ours)  &  p-cloud & \sbest{11.58 $\pm$ 0.08} \\
    \bottomrule
  \end{tabular}}
\end{SCtable}

\begin{table}
 \setlength{\tabcolsep}{3pt}
  \caption{\small FID scores for 3D shape generation. (\emph{lower is better}) All the methods except ours are voxel based.}\vspace{-1em}
  \label{table:fid}
  \centering
  \scalebox{0.75}{
  \begin{tabular}{lcccccccccc}
    \toprule
    Method     & \rotatebox{90}{Dresser} & \rotatebox{90}{Toilet} & \rotatebox{90}{Stand} & \rotatebox{90}{Chair} & \rotatebox{90}{Table} & \rotatebox{90}{Sofa} & \rotatebox{90}{Monitor} & \rotatebox{90}{Bed} & \rotatebox{90}{Bathtub} & \rotatebox{90}{Desk}  \\
    \midrule
    3D-GAN \cite{wu2016learning} \venue{(NIPS'16)}& - & - & - & \sbest{469} & - & 517 & - & - & -& 651 \\
    3D-DesNet \cite{xie2018learning} \venue{(CVPR'18)} & \sbest{414} & 662 & 517 & 490 & 538 & 494 & 511 & 574 & - & - \\
    3D-WINN \cite{huang20193d} \venue{(AAAI'19)} & \best{305} & \sbest{474} & \sbest{456} & \best{225} & \best{220} & \best{151} & \best{181} & \best{222} & \sbest{305} & \best{322}\\
    \midrule
    Spectral-GAN (ours) & 462 & \best{195} & \best{452} & {472} & \sbest{522} & \sbest{180} & \sbest{192} & \sbest{230} & \best{208} & \sbest{354} \\
    \bottomrule
  \end{tabular}}
\end{table}

\noindent \textbf{Quantitative analysis: } To assess the proposed approach quantitatively, we compare the Inception Score (IS) of our network with other voxel-based generative models in Tab.~\ref{table:inception}. In this experiment, we use \cite{qi2016volumetric} as the reference network. IS evaluates a model in terms of both quality and diversity of the generated shapes. Overall, our model demonstrates the second highest performance with a score of $11.58$. To the best of our knowledge,  our work is the first 3D point-cloud GAN to report IS.



We further evaluate our model using Frechet Inception Distance (FID) proposed by Heusel \textit{et al.} \cite{heusel2017gans}, and compare with state-of-the-art. 
IS does not always coincide with human judgement regarding the quality of the generated shapes, as it does not directly capture the similarity between the synthetic and generated shapes. Therefore, FID is used as a complementary measure to evaluate GAN performance. Huang \textit{et al.} \cite{huang20193d} were the first to incorporate FID to 3D GANs, and following them, we also use  \cite{qi2016volumetric} as the reference network. As evident from Table \ref{table:fid}, our results are on-par with state-of-the-art, getting highest scores in three categories: toilet, night stand and bath tub. Interestingly, our Spectral-GAN generally performs better with objects that have curved boundaries, which is a favorable characteristic, as curved boundaries are generally difficult to generate in Euclidean spaces. Note that we convert the point-clouds to meshes before evaluating with both IS and FID.

\noindent \textbf{Comparison with point-cloud generation approaches:}
We use two metrics proposed in Achlioptas \etal \cite{achlioptas2017learning} (i.e., MMD-CD, MMD-ED) to compare the performance of the proposed architecture with other point-cloud generation methods, and display the results in Table \ref{table:pcmeasure}. In this experiment, we use $16$ classes of ShapeNet \cite{yi2016scalable}. As shown, our network gives best results. Intuitively, this suggests that shapes generated by our network have high fidelity compared to the test set. 

\begin{table}
\setlength{\tabcolsep}{4pt}
  \caption{\small Comparison with point-cloud generative models.}\vspace{-1em}
  \label{table:pcmeasure}
  \centering\scalebox{0.75}{
  \begin{tabular}{llcc}
    \toprule
    Method     & Class & MMD-CD & MMD-EMD \\
    \midrule
     r-GAN (dense) \cite{achlioptas2017learning} & & 0.0029 & 0.136 \\
     r-GAN (conv) \cite{achlioptas2017learning} & & 0.0030 & 0.223  \\
     Valsesia \textit{et al.} (no up.) \cite{valsesia2018learning} & Chair & 0.0033  & 0.104 \\
     Valsesia \textit{et al.} (up.) \cite{valsesia2018learning} & & 0.0029 & 0.097 \\
     TreeGAN   \cite{shu20193d} &   & \sbest{0.0016} & 0.101 \\
     Spectral-GAN (ours) &  & \best{0.0012} & \best{0.080}   \\
     
     \midrule
     
      r-GAN (dense) \cite{achlioptas2017learning} &  & 0.0009  & 0.094  \\
     r-GAN (conv) \cite{achlioptas2017learning} &  & 0.0008 & 0.101  \\
     Valsesia \textit{et al.} (no up.) \cite{valsesia2018learning} & Airplane & 0.0010 & 0.102  \\
     Valsesia \textit{et al.} (up.) \cite{valsesia2018learning} &  & 0.0008 & 0.071 \\
     TreeGAN   \cite{shu20193d}& & 0.0004 & 0.068 \\
     Spectral-GAN (ours) & & \best{0.0002} & \best{0.057} \\
     
     \midrule
     
      r-GAN (dense) \cite{achlioptas2017learning} &  & \best{0.0020}  & 0.146 \\
     r-GAN (conv) \cite{achlioptas2017learning} &  & 0.0025  & 0.110  \\
     Valsesia \textit{et al.} (no up.) \cite{valsesia2018learning} & Sofa & \sbest{0.0024} & 0.094 \\
     Valsesia \textit{et al.} (up.) \cite{valsesia2018learning} &  & \best{0.0020} & \sbest{0.083}  \\
     Spectral-GAN (ours) & & \best{0.0020} & \best{0.080} \\
     
     \midrule
     
      r-GAN (dense) \cite{achlioptas2017learning} &  & 0.0021 & 0.155 \\
     TreeGAN   \cite{shu20193d} & All classes   & \sbest{0.0018} & \sbest{0.107}  \\
     Spectral-GAN (w/o backward pass) & & 0.0020 & 0.127\\
     Spectral-GAN (ours) & & \best{0.0015} & \best{0.097} \\
    \bottomrule
  \end{tabular}}
\end{table}

\begin{figure}[!htp]
\centering
\includegraphics[width=1\columnwidth]{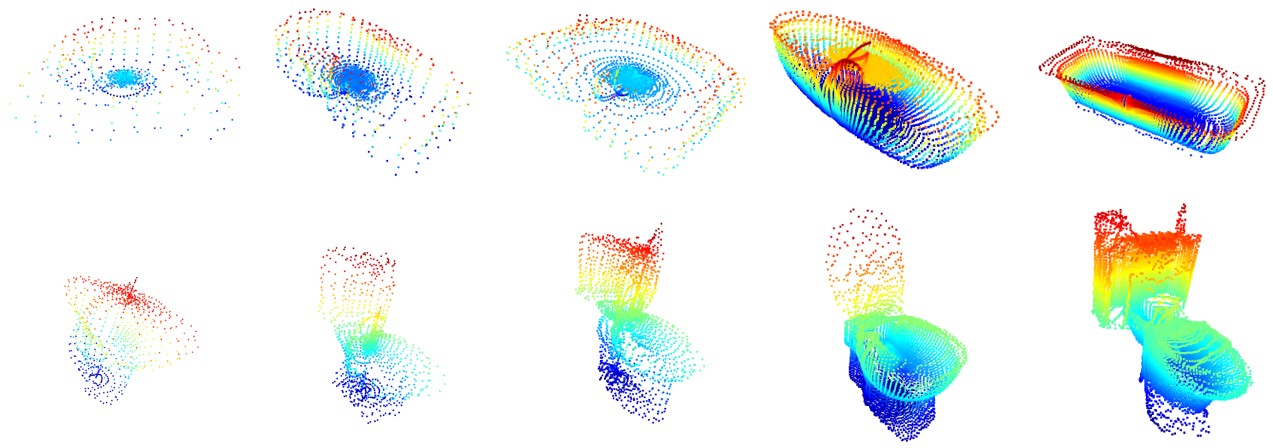}\vspace{-1em}
\caption{\small Scalability of the proposed network with resolution. We obtain increasingly dense resolution by only changing the output layer size in each training phase. Number of points from the left: $30^2, 60^2, 100^2, 150^2$ and $200^2$}.
\label{fig:degree}
\end{figure}

\begin{SCfigure}
\centering
\includegraphics[width=0.5\columnwidth]{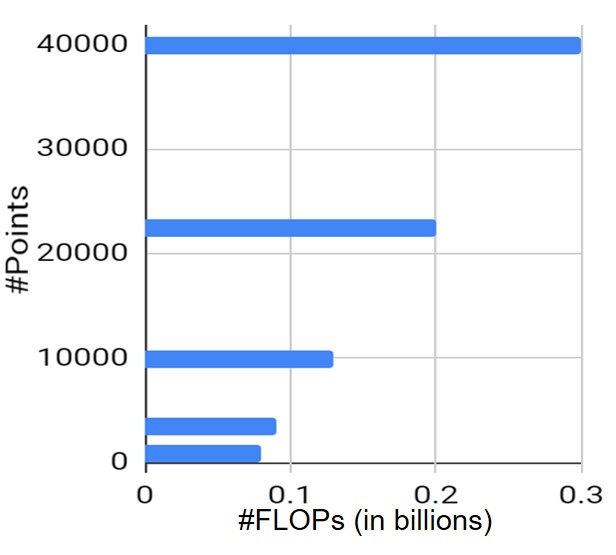}
\caption{\small Spctral GAN can generate high-resolution outputs with minimal computational overhead. We increase resolution approximately $40\times$ while only an increase of $4\times$ FLOPs.}
\label{fig:flops}
\end{SCfigure}

\noindent \textbf{Scalability to high resolutions:} A favorable attribute of our network design is the ability to scale to higher resolutions with minimal changes to the architecture. To verify this, we vary the degree of SMV, and train our model separately for each case. Since the number of points $n$ is  tied to the maximum degree $M$ of SMVs as $n{=} 4M^2$, we obtain samples with different resolutions for each case (see Fig.~\ref{fig:degree}). A key point here is that we only change the output layer size of the generator (according to the length of SMV) to generate point-clouds with different resolutions. Fig.~\ref{fig:flops} illustrates the variation of resolution with the number of FLOPs. Remarkably, we are able to generate high-resolution outputs up to $40,000$ points with only $0.3B$ FLOPs. Another intriguing observation is that our network is able to increase the output resolution by a factor of 40, while the number of FLOPs is only increased by a factor around $4$.

\begin{figure}[tp]
\centering
\includegraphics[width=0.45\textwidth]{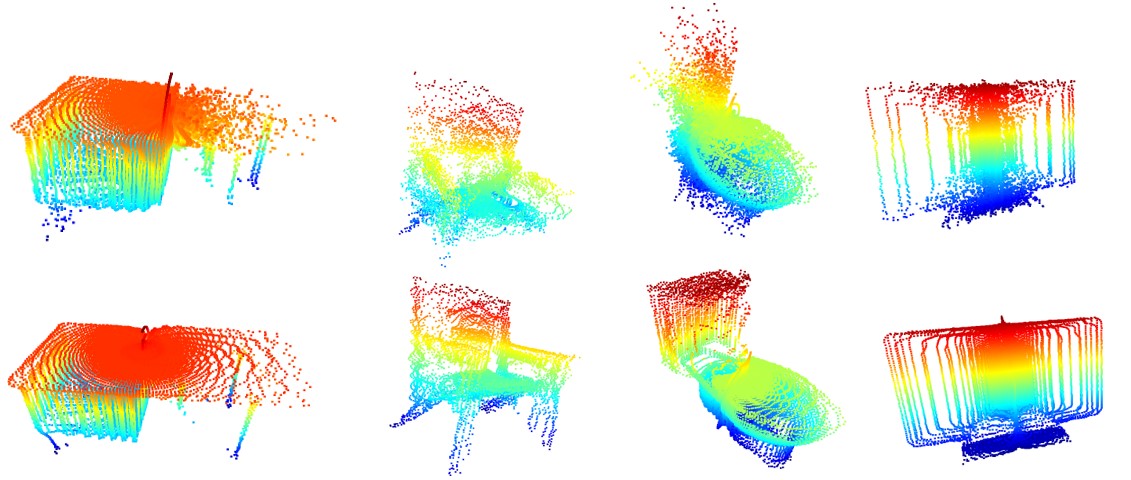}\vspace{-1em}
\caption{\small Effect of backward pass. Top row: samples generated using only forward pass. Bottom row: same samples after passing through both forward and backward pass. Backward pass refines the image by adding more fine details. }
\label{fig:fb}
\end{figure}

\noindent \textbf{Usefulness of  backward pass: } Fig. \ref{fig:fb} illustrates the effect of performing a backward pass. As shown, the forward pass only generates a coarse representation of the shapes without fine details. This is anticipated, since cascaded GANs can only observe the lower frequency portions of SMV in the forward pass. In contrast, the backward pass observes the higher frequency portions, and fine tunes the coarse representation by adding complementary details.

\subsection{Unsupervised 3D Representation Learning}
In this section, we evaluate the representation learning capacity of our discriminator. To this end, we pass relevant SMV frequency bands of 3D point-clouds through trained discriminators, extract the features from the third FC layer, and finally concatenate them to create a feature vector. This feature vector is then fed through a binary SVM classifier and the classification results are obtained as one-against-the-rest. The classification results on ModelNet10 are depicted in Table \ref{table:classification}. As evident, we achieve the highest result with a value of $93.1\%$, which highlights the excellent representation learning capacity of our discriminators.

\subsection{3D reconstruction results}

\begin{SCfigure}[][tp]
\centering
\includegraphics[width=0.35\textwidth]{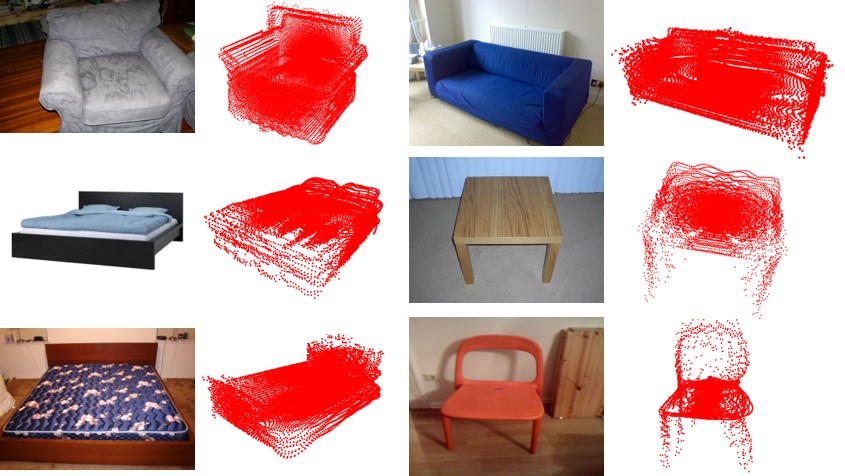}\vspace{0em}
\caption{\small Qualitative results for 3D point-cloud reconstruction from a single image. 
} 
\label{fig:reconstruction}
\end{SCfigure}

In this section, we evaluate the performance of the 3D reconstruction network proposed in Sec. \ref{sec:reconstruction}.
  First, we randomly apply a rotation $R = (R_x,R_y,R_z)$ to each 3D model from the IKEA dataset 15 times, and render the rotated model in front of background images obtained from \cite{xiao2010sun}. Afterwards, we save the rendered images and the corresponding 3D models to create ground-truth image-3D model pairs. The ground truth 3D-models are manually aligned using the Iterative closest point (ICP) algorithm. While applying rotations, we set the constraints $-\frac{\pi}{6}<R_x,R_y<\frac{\pi}{6}$ and $-\pi < R_z<\pi$ and crop the rendered images for the object to be in the center of the images. For the test set, we use the original images provided in the IKEA dataset and test our network on four object classes: chair, sofa, table and bed. We train our model separately for each category and use mean average precision (mAP) to evaluate the performance. In evaluation, we voxelize the generated  and ground truth point-clouds using a $20 {\times} 20 {\times} 20$ voxel grid, and obtain average precision for voxel prediction. The results and illustrative examples are shown in Table \ref{table:pareconstruction} and Fig. \ref{fig:reconstruction}, respectively. As depicted, we surpass state-of-the-art results in sofa and bed categories, while achieving second best results in the table category.
  
\begin{SCtable}[][tp]
  \caption{\small Average precision for 3D point-cloud reconstruction from single image. The point-clouds are voxelized before obtaining the score.}
  \label{table:pareconstruction}
  \centering \setlength{\tabcolsep}{5pt}
  \scalebox{0.65}{
  \begin{tabular}{lcccc}
    \toprule
    Method     & Chair & Sofa & Bed & Table\\
    \midrule
     AlexNet-fc8 \cite{girdhar2016learning} & 20.4 & 38.8 & 29.5 & 16.0 \\
     AlexNet-conv4 \cite{girdhar2016learning} & 31.4 & 69.3 & 38.2 & 19.1 \\
     T-L network \cite{girdhar2016learning} & 32.9 & 71.7 & 56.3 & 23.3 \\
     3D-VAE-GAN \cite{wu2016learning} & 47.2 & \sbest{78.8} & 63.2 & 42.3 \\
     VAE-IWGAN \cite{smith2017improved} & \best{49.3} & 68.0 & 65.7 & 52.2\\
     PrimtiveGAN \cite{khan2019unsupervised} & \sbest{47.5}  & 77.1 & \sbest{68.4} & \best{60.0} \\
     \midrule
     Spectral-GAN (ours) & 42.3 & \best{81.2} & \best{71.4} & \sbest{48.3} \\
    \bottomrule
  \end{tabular}}
\end{SCtable}

\section{Conclusion}
We propose a generative model for 3D point-clouds that operates in the spectral-domain. In contrast to previous methods that operate in the spatial-domain, our approach provides a structured way to deal with the inherent redundancy and irregularity of point-clouds. We demonstrate that our model generates sound 3D outputs, can scale to high-dimensional outputs and learns discriminative features in an unsupervised manner. Further, it can be used for 3D reconstruction task.

{\small
\bibliographystyle{ieee_fullname}
\bibliography{egbib}
}

\newpage
\appendix
\onecolumn

\setcounter{figure}{0}
\setcounter{table}{0}

\begin{center}
{\Large\bf Supplementary Materials\\[0.5em]
Spectral GAN for High-Resolution 3D Point-cloud Generation}
\end{center}
\vspace{1em}

This supplementary material provides details about the sampling and reconstruction procedure, compares our work with the previous cascaded generative designs and provides additional qualitative and quantitative results. 

\section{Sampling and reconstruction}
A key attribute of any sampling theorem is the minimum number of sample points required to accurately represent a band-limited function in a particular space. Several such sampling theorems have been proposed to represent a signal with finite energy in $\mathbb{S}^2$, whereas a most popular choice is the Driscoll and Healy’s (DH) theorem proposed by Driscoll \textit{et al.} \cite{driscoll1994computing}, which we also use in our work.

According to DH theorem, to accurately represent a signal on $\mathbb{S}^2$ using spherical harmonic moments band-limited at degree $M$, $4M^2$ equiangular sampled points are needed. For all the main experiments in this work, we choose $M = 100$ and obtain an equally sampled $200 \times 200$ grid in each $\theta$ and $\phi$ directions, where $0 \leq \theta \leq \pi$ and $0 \leq \phi \leq 2\pi$. However, as mentioned in Sec. \ref{equ:expansion}, spherical harmonics can represent only polar 3D shapes, which can result in less visually pleasing spatial representations of non-polar shapes. To overcome this obstacle, we follow the following sampling procedure.

\begin{figure}[!hb]
\centering
\includegraphics[width=0.5\textwidth]{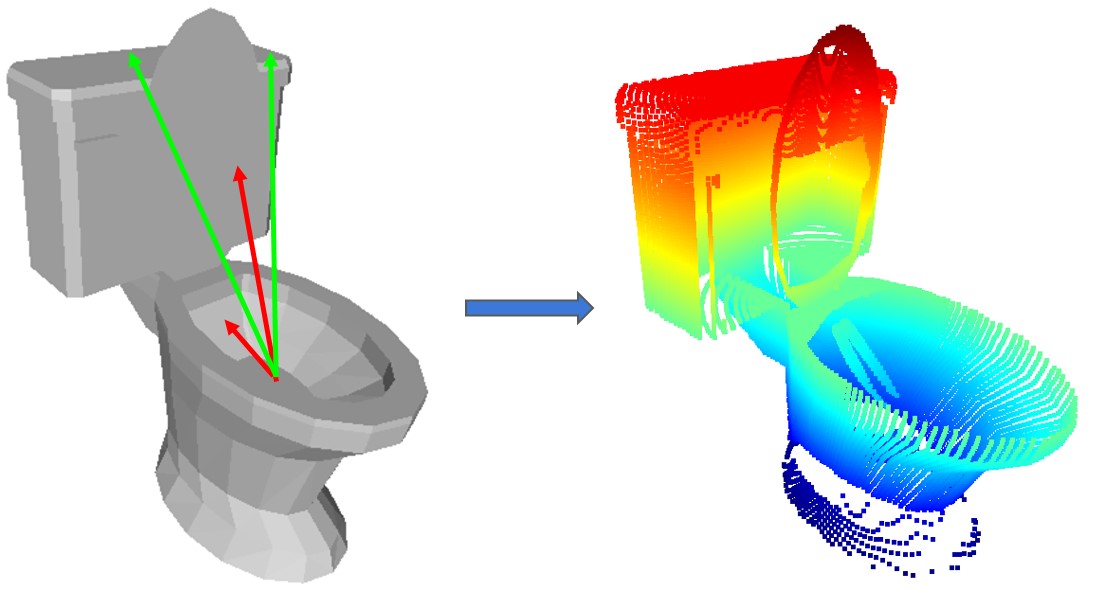}
\caption{Illustration of the sampling procedure. Red arrows and green arrows demonstrate first stage and second stage sampling, respectively.}
\label{fig:sampling}
\end{figure}

First, we scale the 3D mesh to fit inside the unit ball $\mathbb{B}^3$, and cast rays from the centroid of the shape to outward direction, and take the first hit locations of the rays with a face as a sample point. In the first stage, we sample $200 \times 100$ such equiangular points in a $200 \times 100$ grid, sampled in $\theta$ and $\phi$ directions respectively, where $0 \leq \theta \leq \pi$ and $0 \leq \phi \leq 2\pi$. In the second stage, we rotate the casted rays in $\phi$ direction, by an amount of $\frac{\pi}{99}$, and obtain the last hit locations of the each ray with a face of the 3d shape as a sample point. Union of these two sampling sets provide a more visually pleasing point-cloud for non-polar 3D shapes. This procedure is illustrated in Fig. \ref{fig:sampling}.

\section{Literature on cascaded generative designs:}
Denton \textit{et al.} \cite{denton2015deep} proposed a cascaded GAN architecture for 2D image generation. Similar to our work, they also use a series of conditional GANs which are conditioned upon one another. These GANs generate image representations in a Laplacian pyramid framework to create increasingly refined images. Instead of generating images directly in the spatial domain, these generative models specialize in generating a specific residual image, according to the corresponding stage of the Laplacian pyramid, which are finally combined together to produce a high quality image. This is analogous to our work, where our generators generate a specific frequency portion of SMVs, which are finally combined together to obtain the full representation. Other recent works also employ cascaded generative architectures to improve image quality e.g., \cite{wang2018high} use a combination of generators operating on low and high resolution domains, \cite{Wang_ssganECCV2016} separately train generative models to learn style and structure components, \cite{Zhang_2017_ICCV} progressively adds photorealistic details in low-resolution generated images. The conditional stacked GAN architecture of Huang \etal \cite{Huang_2017_CVPR} is particularly close to ours, that feeds onto previous generators output and new latent vectors to create novel images. Finally, the seminal SinGAN \cite{Shaham_2019_ICCV} approach designs a pyramid of coarse-to-fine generators that can be trained on a single image. However, as opposed to current work, all above efforts operate in the spatial domain and have no concrete definition of spectral bands.

\section{Computational complexity analysis}

A key feature of our network is its high computational efficiency despite being a cascaded design. Since the target is a 1-D structured vector, the generators are allowed to have a shallow architecture, which decreases the total number of FLOPs during operation. Table \ref{table:complexity} compares the our model complexity against the state-of-the-art models. We achieve the best performance in terms of MMD-CD and MMD-EMD while having the lowest model complexity. Experiments are conducted for inference with 20 batch size.

\begin{figure}
\centering
\includegraphics[width=1.0\textwidth]{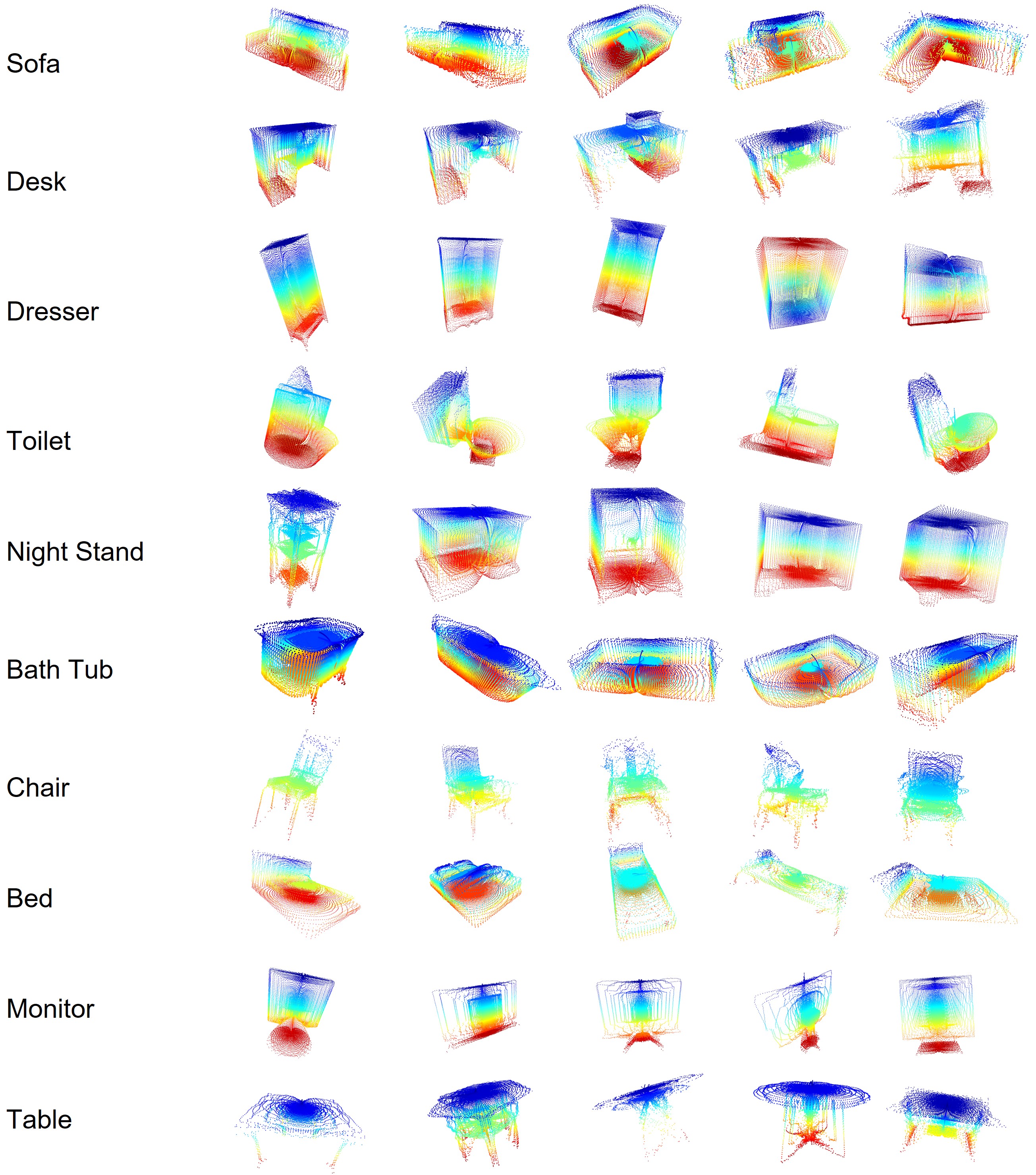}
\caption{Qualitative results: generated point clouds for each class.}
\label{fig:samples}
\end{figure}

\begin{figure}
\centering
\includegraphics[width=1.0\textwidth]{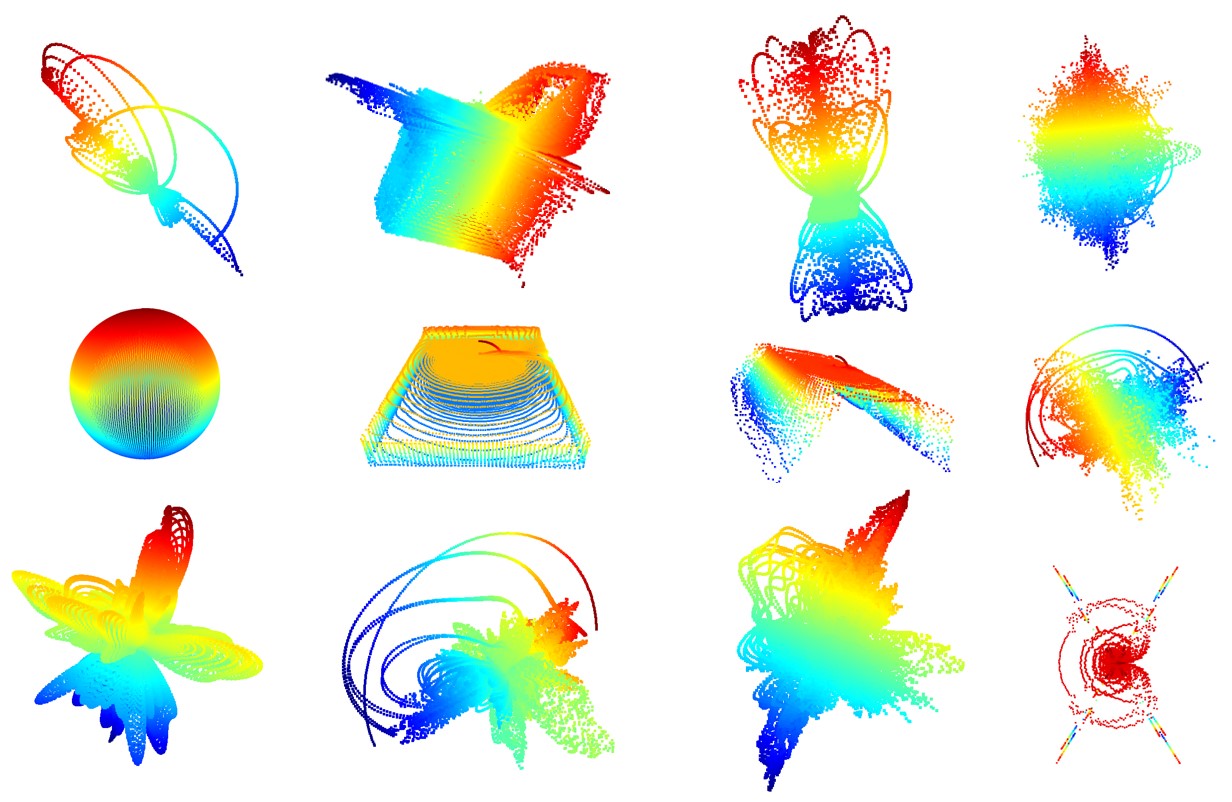}
\caption{ Our network tends to generate weird artifacts among plausible samples, when trained without the spatial domain regularizer, since small variations in spectral domain cause significant variations in spatial domain. A few such examples are illustrated here. These artifacts are effectively suppressed by our spatial domain regularizer.}
\label{fig:samples}
\end{figure}

\begin{table}[!htp]
  \caption{\small Model complexity comparison with point-cloud generative models (inference). We achieve the best performance while having the lowest complexity. ($\downarrow$ denotes lower is better, $\uparrow$ denotes higher is better)}
  \label{table:complexity}
  \centering\scalebox{0.75}{
  \begin{tabular}{llcccc}
    \toprule
    Method     &  & MMD-CD ($\downarrow$) & MMD-EMD ($\downarrow$) & \#FLOPs ($\downarrow$) & \#Points ($\uparrow$)  \\
    \midrule
     r-GAN (dense) \cite{achlioptas2017learning} & & 0.0029 & 0.136 & 0.1B & 2048\\
     Valsesia \textit{et al.} (up.) \cite{valsesia2018learning} & Chair & 0.0029 & 0.097 & 304B & 2048 \\
     Spectral-GAN (ours) &  & \best{0.0012} & \best{0.080} & \best{0.09B} & \best{3600}   \\
     
     \midrule
     
      r-GAN (dense) \cite{achlioptas2017learning} &  & 0.0009  & 0.094 & 0.1B & 2048  \\

     Valsesia \textit{et al.} (up.) \cite{valsesia2018learning} & Airplane  & 0.0008 & 0.071 & 304B & 2048 \\
     Spectral-GAN (ours) & & \best{0.0002} & \best{0.057} & \best{0.09B} & \best{3600} \\
     
     \midrule
     
      r-GAN (dense) \cite{achlioptas2017learning} &  & \best{0.0020}  & 0.146 & 0.1B & 2048 \\
     Valsesia \textit{et al.} (up.) \cite{valsesia2018learning} & Sofa & \best{0.0020} & 0.083 & 304B & 2048  \\
     Spectral-GAN (ours) & & \best{0.0020} & \best{0.080} & \best{0.09B} & \best{3600} \\
     
     \midrule
     
      r-GAN (dense) \cite{achlioptas2017learning} & All classes  & 0.0021 & 0.155 & 0.1B & 2048 \\
     Spectral-GAN (ours) & & \best{0.0015} & \best{0.097} & \best{0.09B} & \best{3600} \\
    \bottomrule
  \end{tabular}}
\end{table}

\end{document}